\documentclass[twoside]{article}

\usepackage[arxiv]{aistats2023}

\usepackage[round]{natbib}

\usepackage{amsmath,amsfonts,bm}

\def\figref#1{figure~\ref{#1}}

\def\eqref#1{equation~\ref{#1}}
\def\Eqref#1{Equation~\ref{#1}}

\def\1{\bm{1}}

\DeclareMathAlphabet{\mathsfit}{\encodingdefault}{\sfdefault}{m}{sl}
\SetMathAlphabet{\mathsfit}{bold}{\encodingdefault}{\sfdefault}{bx}{n}

\newcommand{\KL}{D_{\mathrm{KL}}}

\usepackage[utf8]{inputenc} 
\usepackage[T1]{fontenc}    
\usepackage{hyperref}       
\usepackage{url}            
\usepackage{booktabs}       
\usepackage{amsfonts}       
\usepackage{nicefrac}       
\usepackage{microtype}      

\usepackage{array}

\usepackage{amsmath}
\usepackage{amssymb}
\usepackage{commath}
\usepackage{bbm}

\usepackage{booktabs}
\usepackage{multirow}
\usepackage{pgfplots}
\pgfplotsset{compat=1.16}

\newcommand{\lam}{\lambda}
\newcommand{\lamt}{{\lam_\text{T}}}
\newcommand{\Ot}{{O_\text{T}}}
\newcommand{\Lam}{\Lambda}
\newcommand{\reals}{\mathbb{R}}

\newcommand{\pt}{p_\text{T}}
\newcommand{\pp}{p_\lam}
\newcommand{\qq}{q_\nu}

\newcommand{\get}[1]{\overline{#1}}

\newcommand{\half}{\tfrac{1}{2}}

\newcommand{\tr}[1]{{#1}^\mathsf{T}}

\newcommand{\ones}{\mathbbm{1}}

\newcolumntype{M}[1]{>{\centering\arraybackslash}m{#1}}
\newcolumntype{C}[1]{>{\centering\arraybackslash}p{#1}}
\usepackage[justification=centering]{caption}

\let\KL\relax
\DeclareMathOperator{\KL}{KL}

\newcommand{\divergence}[2]{(#1\,\|\,#2)}
\newcommand{\KLD}[2]{\KL\divergence{#1}{#2}}

\newcommand{\sref}[1]{\S\ref{#1}}
\newcommand{\tabref}[1]{Table~\ref{#1}}
\renewcommand{\figref}[1]{Figure~\ref{#1}}
\renewcommand{\Eqref}[1]{(\ref{#1})}

\usepackage[symbol]{footmisc}

\begin{document}

\twocolumn[

\aistatstitle{Learning the joint distribution of two sequences using little or no paired data}

\aistatsauthor{
  Soroosh Mariooryad
  \footnotemark[1]
  \footnotemark[3]
  \And Matt Shannon
  \footnotemark[2]
  \footnotemark[3]
  \And Siyuan Ma
  \And Tom Bagby
}
\aistatsauthor{
  David Kao
  \And Daisy Stanton
  \And Eric Battenberg
  \And RJ Skerry-Ryan
}

\runningauthor{
Soroosh Mariooryad, Matt Shannon, {\textit{et al.}}
}

\aistatsaddress{Google, Mountain View, California, USA}

]

\footnotetext[1]{soroosh@google.com}
\footnotetext[2]{mattshannon@google.com}
\footnotetext[3]{These authors contributed equally.}

\begin{abstract}
We present a noisy channel generative model of two sequences, for example text and speech, which enables uncovering the association between the two modalities when limited paired data is available. To address the intractability of the exact model under a realistic data setup, we propose a variational inference approximation. To train this variational model with categorical data, we propose a KL encoder loss approach which has connections to the wake-sleep algorithm. Identifying the joint or conditional distributions by only observing unpaired samples from the marginals is only possible under certain conditions in the data distribution and we discuss under what type of conditional independence assumptions that might be achieved, which guides the architecture designs. Experimental results show that even tiny amount of paired data ($5$ minutes) is sufficient to learn to relate the two modalities (graphemes and phonemes here) when a massive amount of unpaired data is available, paving the path to adopting this principled approach for all seq2seq models in low data resource regimes.
\end{abstract}

\section{INTRODUCTION}
\label{sec:intro}
Learning the joint or conditional distribution of two sequences appears in many machine learning applications such as automatic speech recognition (ASR),
text-to-speech (TTS) synthesis, machine translation, optical character recognition, and text summarization. When there is limited or no paired data, we would like to learn these distributions from large amounts of unpaired data. While the proposed approach is generally applicable to all these seq2seq problems, we ground the discussions primarily on text and speech for ASR and TTS applications.

A classical ASR approach treats the process of generating speech as a noisy channel. In this framing, text is drawn from some distribution and statistically
transformed into speech audio; the speech recognition task is then to invert this
generative model to infer the text most likely to have given rise to a given speech waveform.
This generative model of speech was historically successful \citep{baker1975stochastic,jelinek1976continuous,rabiner1989tutorial},
but has been superseded in modern discriminative systems by directly modeling the conditional distribution of text, given speech \citep{graves2006connectionist,amodei2016deep}.
The direct approach has the advantage of allowing limited modeling power to be solely devoted
to the task of interest, whereas the generative one can be extremely sensitive to faulty
assumptions in the speech audio model despite the fact that this is not the primary object of
interest.
However the generative approach allows learning in a principled way from untranscribed
speech audio, something fundamentally impossible in the direct approach.
We explore a noisy channel joint model of text and speech for learning from
a corpus consisting of relatively large amounts of text-only and speech-only data, but
little or no parallel (text, speech) data.
We cope with the sensitivity of generative modeling to faulty modeling assumptions by trying to
make the generative model as accurate as possible, and cope with the intractable inference
problem using a approximate variational posterior of text given speech. An analogous formulation in the other direction can be adopted for
TTS models.

Similar to discrete latent variable models, when our variational approach infers a discrete quantity (e.g., text), the typical
stochastic gradient variational Bayes approach \citep{kingma2013auto} is not applicable. We propose a new optimization procedure applicable to the discrete case (\sref{sec:kl-encoder-loss}).

The large body of work on leveraging speech-only and text-only data resources to build ASR and TTS systems rely on the close connection between the two modalities. However, the feasibility and necessary conditions to doing so is not
well understood theoretically. Here we formalize the problem as identifiying the joint text and speech distribution by only observing its marginal samples, and
try to make a dent in this problem. Particularly, we hypothesize that certain conditional independence assumptions are required to recover the joint and present
preliminarily identifiability proofs for a constrained variation of the problem.

Our core contributions are:
\begin{itemize}
    \item Describing the noisy channel model of joint distribution in \sref{sec:model}.
    \item Deriving a variational noisy channel model in \sref{sec:model}.
    \item The KL encoder loss to train discrete latent variable models in \sref{sec:kl-encoder-loss}.
    \item Preliminary discussions on identifiability given no paired data in \sref{sec:ident}.
    \item Experiments demonstrating breaking a grapheme substitution cipher with no paired data, and learning to relate speech and text with limited paired data in \sref{sec:experiments}.
\end{itemize}

\section{MODEL}
\label{sec:model}
In this section we describe our joint model of two sequences and how to train it.
The two sequences $x = [x_s]_{s=0}^{S-1}$ and $y = [y_t]_{t=0}^{T-1}$ may be different lengths ($S \neq T$)
and may each have discrete or continuous values.
For example, the first sequence $x$ might be text consisting of a sequence of graphemes and the
second sequence $y$ a sequence of mel spectrogram frames in an application related to speech
recognition and synthesis, or $x$ might be text and $y$ a sequence of image patches corresponding to
printed characters in an application related to optical character recognition.

We assume we have a mix of paired and unpaired data.
Specifically we assume the corpus is generated by repeatedly and independently sampling a sequence pair
$(x, y)$ from the \emph{true distribution} (or \emph{data distribution}) $\pt(x, y)$
and then keeping only $x$ with probability $\alpha$, only $y$ with probability $\beta$, or
both $x$ and $y$ with probability $\gamma$, where $\alpha + \beta + \gamma = 1$.
We refer to $\gamma$ as the \emph{paired fraction}.
We are interested in the regime $\gamma \ll 1$, including the extreme case $\gamma = 0$ where
there is no paired data.

One advantages of generative modeling is that it provides a principled way to use unpaired data
during parameter estimation \citep{cooper1970asymptotic}.
The model $\pp(x, y)$ defines a joint distribution over the two sequences $x$ and $y$ with parameters 
$\lam$, which in turn defines marginals $\pp(x)$ and $\pp(y)$.
If the marginals are tractable then we may estimate $\lam$ by minimizing the cross-entropy
\begin{equation}
  \label{eq:loss}
  -\sum_u \pt(u) \log\pp(u)
\end{equation}
where $u$ is ``whatever is observed'' for a given example, be that $x$ or $y$ or $(x, y)$.
In practice the expectation over $\pt(u)$ is replaced with samples from the training corpus
yielding a form of maximum likelihood estimation.
We may write \Eqref{eq:loss} concisely as the KL divergence $\KLD{\pt(u)}{\pp(u)}$ with the
understanding that the unknown but irrelevant additive constant $\sum_u \pt(u) \log\pt(u)$ is not
computed in practice.
This KL divergence can in turn be written as
\begin{equation}
  \label{eq:loss-decomp}
  \begin{split}
  \KLD{\pt(u)}{\pp(u)}
    &= \alpha \KLD{\pt(x)}{\pp(x)}
  \\
    &\quad + \beta \KLD{\pt(y)}{\pp(y)}
  \\
    &\quad + \gamma \KLD{\pt(x, y)}{\pp(x, y)}
  \end{split}
\end{equation}
Thus this loss incentivizes the model to match both the marginal and joint distributions
of the data.

The generative model used in this work is a form of noisy channel model.
We factorize $\pp(x, y)$ in terms of a \emph{prior} $\pp(x)$ and \emph{decoder} $\pp(y | x)$.
We typically use recurrent autoregressive models with step-by-step end-of-sequence decisions for both the
prior and decoder, using attention to incorporate the conditioning information $x$ for the decoder
(see \sref{sec:setup}).
The noisy channel model allows directly computing $\pp(x)$ and $\pp(x, y)$.
The marginal $\pp(y)$ is tractable for simple models such as a Markovian prior and decoder.
When the marginal is not tractable, we introduce a \emph{variational posterior} (or \emph{encoder})
$\qq(x | y)$ and replace $\KLD{\pt(y)}{\pp(y)}$ in \Eqref{eq:loss-decomp} with the upper bound
\begin{equation}
  \label{eq:klubo}
  \KLD{\pt(y) \qq(x | y)}{\pp(x, y)}
\end{equation}
This is the conventional negative \emph{evidence lower bound objective (ELBO)} \citep{beal2000variational}
up to a constant.
In contrast to variational latent variable models such as \emph{variational autoencoders (VAEs)}
\citep{kingma2013auto},
here the space modeled by the prior and variational posterior is grounded by observed data.
In the case $x$ is text and $y$ is speech, $\pp(y | x)$ is a TTS model and $\qq(x | y)$ is an ASR model.

\subsection{KL encoder loss}
\label{sec:kl-encoder-loss}
To cope with discrete-valued $x$, we propose a novel variant of the wake-sleep algorithm \citep{hinton1995wake}.
In this section we describe this approach.

We first review why discrete $x$ is more challenging than continuous $x$. The expression \Eqref{eq:klubo}
involves an expectation over $\qq(x | y)$.
If $x$ is a sequence of continuous values of known length then this expectation can be reparameterized,
allowing low variance finite sample approximations to the gradient of this term with respect to $\nu$
\citep{kingma2013auto}.
However if $x$ is discrete this is not possible.
There have been many alternative methods proposed to compute finite
sample approximations to the gradient, including REINFORCE \citep{williams1992simple},
RELAX \citep{grathwohl2017backpropagation} and many others.
In our application, this challenge applies even if $x$ has continuous values, since the length of $x$ is
unknown and discrete.

We solve this problem by modifying the loss used to train the variational posterior.
Instead of minimizing $\KLD{\pt(y) \qq(x | y)}{\pp(x, y)}$ with respect to the encoder parameters $\nu$,
we instead minimize $\KLD{\pp(x, y)}{\pt(y) \qq(x | y)}$ with respect to $\nu$.
We continue to train the generative model parameters $\lam$ as before.
This training procedure is reminiscent of the wake-sleep algorithm \citep{hinton1995wake},
where the $\lam$ updates and $\nu$ updates correspond to the wake phase and sleep phase respectively.
We refer to this approach as \emph{KL encoder loss} training since the variational posterior appears in the
right ``KL'' argument to the KL divergence, as opposed to the conventional ELBO for which the variational posterior
appears in the left ``reverse KL'' argument to the KL divergence.
The conventional ELBO and KL encoder loss have the same non-parametric optimal variational posterior, namely
$\hat{q}(x | y) = \pp(x | y)$.
The two approaches place different computational demands on $\qq$ and $\pp$.
The conventional approach requires tractable reparameterized samples and log prob computations for $\qq(x | y)$
and tractable log prob computations for $\pp(x, y)$.
The KL encoder loss approach requires tractable log prob computations for $\qq(x | y)$ and tractable samples
for $\pp(x, y)$.

The use of different objectives for different parts of the model is reminiscent of GAN training,
but note that here the losses are cooperative rather than adversarial, in the sense that making the
variational posterior optimal improves both the variational loss and the generative loss, whereas making the
critic optimal in classic GAN training makes the generator loss worse.
Nevertheless, there is no guarantee that the training dynamics of the (generative, variational) system
are convergent in general.%
\footnote{%
  If the learning rate used for the generative parameters $\lam$ is set sufficiently small relative to
  the learning rate used for the variational parameters $\nu$, and the variational posterior is
  sufficiently flexible, then the variational posterior is able to remain essentially optimal
  throughout training and so the training dynamics are effectively just gradient descent on
  \Eqref{eq:loss-decomp} with respect to $\lam$, which has well-behaved training dynamics.
}

\subsection{Model training summary}
\label{sec:training-summary}
We now summarize our training procedure.
The loss $l^\text{gen}$ used to learn the parameters $\lam$ of the generative model
$\pp(x, y) = \pp(x) \pp(y | x)$ and loss $l^\text{var}$ used to learn the parameters $\nu$ of the
variational posterior $\qq(x | y)$ are
\begin{align}
  \begin{split}
  \label{eq:loss-generative}
  l^\text{gen}_{\lam; \nu}
    &= -\alpha \sum_x \pt(x) \log \pp(x)
  \\
    &\quad - \beta \sum_{x, y} \pt(y) \qq(x | y) \left[
      \log\pp(x, y) - \log\qq(x | y)
    \right]
  \\
    &\quad - \gamma \sum_{x, y} \pt(x, y) \log\pp(x, y)
  \end{split}
\\
  \label{eq:loss-variational}
  l^\text{var}_{\nu; \lam}
    &= -\sum_{x, y} \pp(x, y) \log\qq(x | y)
\end{align}
In practice each loss is approximated with a stochastic minibatch approximation based
on the training corpus in the natural way, using the $\alpha$ term for examples where
only $x$ is observed, the $\beta$ term for examples where only $y$ is observed, and
the $\gamma$ term for examples where both $x$ and $y$ are observed.
We perform simultaneous gradient descent on $(\lam, \nu)$ based on the gradients
$(\partial l^\text{gen}_{\lam; \nu} / \partial \lam, \partial l^\text{var}_{\nu; \lam} / \partial \nu)$.

We find three experimental tricks helpful for training.
Firstly, samples from autoregressive models can suffer from small errors compounding
over time, particularly when trained with maximum likelihood estimation / KL. This only weakly
penalizes unrealistic next-step samples because KL is a ``covering'' rather than ``mode-seeking'' divergence
\citep[Section 10.1.2]{bishop2006pattern}.
A commonly used trick for both non-autoregressive \citep{parmar2018image,kingma2018glow}
and autoregressive \citep{weiss2021wave} models is to adjust the temperature of the distribution.
The prior, decoder and variational posterior are all trained with KL, and we apply
temperature adjustment when sampling from these models during both training and decoding.
For example, for the variational posterior we recursively sample from
$\tfrac{1}{Z_\nu(x_{0:t-1}, y)} (\qq(x_t | x_{0:t-1}, y))^{\frac{1}{T}}$ instead of
$\qq(x_t | x_{0:t-1}, y)$.
Typically $T = 0.5$.

Secondly, at random initialization the generative model and variational posterior are both very
suboptimal, and the noisy gradients from the $\beta$ term of $l^\text{gen}$
may swamp the small but consistent signal from the paired data $\gamma$ term when training the decoder.
To alleviate this, we pre-train with the $\beta$ term omitted from $l^\text{gen}$, effectively
ignoring the $y$-only data.
Finally, we optionally ignore the ELBO term throughout training when updating the prior $\pp(x)$.
In the regime where $\alpha$ is small this could prevent the model learning important information
about $\pt(x)$ present in the $y$-only data, but in the regime we consider here where there is
plenty of $x$-only data, it slightly helps to stabilize training.

\section{IDENTIFIABILITY}
\label{sec:ident}
In this section we discuss the challenges that exist when little or no paired data is available.
We mainly focus on the case of no paired data.

We first define identifiability given no paired data.
We say a generative model $\pp(x, y)$ is \emph{identifiable given no paired data} if
matching the marginals implies matching the joint,
that is
if $\pp(x) = p_\lamt(x)$ for all sequences $x$ and $\pp(y) = p_\lamt(y)$ for all sequences $y$
implies $\pp(x, y) = p_\lamt(x, y)$ for all sequences $x$ and $y$.
We do not require $\lam = \lamt$.
We may think of $\lamt$ as the true parameters and $\lam$ as the model parameters being learned.

Even in the case where the model is identifiable, local optima may be a substantial impediment
to learning.
These local optima are quite a generic feature of learning from little or no paired data.
For example, if we set $\alpha = \gamma = 0$ in \Eqref{eq:loss-decomp} then $x$ becomes a latent
variable, and so the loss is invariant under permutations of the categories or dimensions used for $x$.
This symmetry over permutations means that it is impossible even in principle to recover the true mapping between $x$ and $y$.
Since \Eqref{eq:loss-decomp} is continuous in $(\alpha, \beta, \gamma)$, small $\alpha$ and $\gamma$ values
will have multiple spurious local optima as the remnants of the spurious global optima which exist at
$\alpha = \gamma = 0$.

We now discuss the need to restrict the power of the decoder.
If the decoder $\pp(y | x)$ is very flexible then it may be possible for it to completely ignore $x$
yet still obtain a perfect marginal $\pp(y) = \pt(y)$.
Clearly this learns nothing about the true mapping between $x$ and $y$.
We therefore restrict the power of the decoder so that it is forced to use $x$.
In contrast we try to make the prior and variational posterior as flexible as possible in order to
ensure accurate modeling and as tight a variational bound as possible.

\subsection{Time locality}
One widely applicable way to limit decoder power is by assuming \emph{time locality}.
We say a decoder has strict time locality if the overall probability can be written as a product
of time-local factors
\begin{equation}
  \label{eq:time-local}
  \pp(y | x) = \prod_{t=0}^{T-1} f_t(y_{t-L:t+L}, x_{\get{s}(t)-K:\get{s}(t)+K})
\end{equation}
where \emph{time constants} $K, L \in \mathbb{Z}_{\geq 0}$ and
$\get{s} : \{0, \ldots, T - 1\} \to \{0, \ldots, S - 1\}$ is a function aligning each position in $y$
to a position in $x$.
For example, the decoder $\pp(y | x) = \prod_t \pp(y_t | x_{\get{s}(t)})$ which is independent over time
and for which each $y_t$ only depends on a single $x_s$ is strictly time local with $K = L = 0$.
We refer to a decoder as time local if \Eqref{eq:time-local} holds approximately.
If we assume that the true marginal $\pt(y)$ has long-range correlations which mean it is either not time local,
or is time local with time constant much greater than $L$, then the only way for the model as a whole
to capture these correlations across time in its marginal $\pp(y)$ is to induce them from corresponding
correlations across time in $x$.
This provides the generative model with an incentive to uncover how $x$ maps to $y$.

Time locality is an intuitively reasonable assumption in many seq2seq problems such as
speech recognition and synthesis, optical character recognition and machine translation (with non-monotonic
$\get{s}$).
It thus forms a promising middle ground as a weak enough assumption to have wide applicability but a
strong enough assumption to potentially support learning the joint with little or no paired data.

\subsection{A worked example of identifiability}
\label{sec:ident-example}
To help guide our intuition surrounding identifiability in more complicated cases, we now examine
identifiability in the case of a Markovian prior and time-independent and time-synchronous decoder
with no paired data available.
In this case $\pp(x, y) = \prod_t \pp(x_t | x_{t-1}) \pp(y_t | x_t)$.
Let $B_{i j} = \pp(x_0=i, x_1=j)$, $O_{i p} = \pp(y_t=p | x_t=i)$, $D_{p q} = \pp(y_0=p, y_1=q)$ and
$\ones$ be a vector of ones.
We assume that the prior is a stationary distribution, that is $B \ones = \tr{B} \ones = b$ and that
$b_i > 0$ for all $i$.
The prior is easy to learn from unpaired data, and so we assume $\pt(x_0=i, x_1=j) = B_{i j}$.
Let $C_{p q} = \pt(y_0=p, y_1=q)$ and $c = C \ones = \tr{C} \ones$.
In this case we can conveniently express the relationship between the $y$ marginals and $x$ marginals
as a matrix multiplication $D = \tr{O} B O$.

We first consider the case where $O$ is a permutation matrix, corresponding to a \emph{substitution cipher}.
We assume $x$ is English text represented as a series of graphemes.
For example, the ciphertext $y$ might be \texttt{wi jtvwjpvwjbhjwi jgvw}, corresponding to some
English plaintext $x$.
It is well-known that this simple cipher can be broken by frequency analysis, by tabulating the frequency
of grapheme n-grams in the ciphertext and looking for grapheme n-grams with similar frequencies in
conventional English text.
We may codify this by considering the singular value decompositions of $B$ and $C$.
We show in \sref{sec:ident-more} that as long as the singular values of $B$ are distinct and non-zero
then we can completely recover $O$ and have identifiability given no paired data.
The plaintext above is \texttt{the cat sat on the mat}.

Secondly we consider the case where $O$ is not restricted to be a permutation matrix but where the
$x$ and $y$ alphabets both have size two, say $x_s, y_t \in \{0, 1\}$.
Since $O \ones = \ones$, there are only two degrees of freedom in $O$, say
\begin{equation}
  O = \left[
    \begin{matrix}
      \eta & 1 - \eta
      \\
      \zeta & 1 - \zeta
    \end{matrix}
  \right]
\end{equation}
We first consider cases where we do not have identifiability, which may be particularly helpful for building
general intuition.
The first degenerate case is where $B$ is low rank, that is $B = b \tr{b}$ and $x_0$ and $x_1$ are independent.
In this case $O$ is never identifiable, since any $(\eta, \zeta)$ on the line $\eta b_0 + \zeta b_1 = c_0$
results in the same unigram marginal $\pp(y_0)$ and so the same overall marginal $\pp(y)$ due to independence
over time.
This is a simple example of needing correlations over time in $x$ that are longer than the decoder can
model on its own in order to have identifiability.
The second degenerate case we consider is where $b_0 = \half$.
In this case
\begin{equation}
  B = \left[
    \begin{matrix}
      B_{0 0} & B_{0 1}
      \\
      B_{0 1} & B_{0 0}
    \end{matrix}
  \right]
\end{equation}
This obeys the symmetry that swapping $0$s and $1$s does not change the probability of a sequence under
the prior.
Intuitively this means we have no way to distinguish which $x$ symbol maps to a given $y$ symbol, just
like in the case where $x$ is a latent variable which is never observed.
Formally $(\eta, \zeta)$ and $(\zeta, \eta)$ result in the same marginal $\pp(y)$ for all $y$.
Technically we do still have identifiability if $\eta = \zeta$, but this case is practically uninteresting
because it means $x$ and $y$ are completely independent.
Otherwise we do not have identifiability when $b_0 = \half$.
By considering sequences of length two and three, we show in \sref{sec:ident-more} that if $B$ is full rank
and $b_0 \neq \half$ then we do have identifiability given no paired data.
The general pattern in this simple case is that the time locality assumption is sufficient to ensure
identifiability unless the marginal $\pp(x)$ obeys one of a finite list of a specific symmetries that
make identification impossible.

\section{EXPERIMENTS}
\label{sec:experiments}
\definecolor{bblue}{HTML}{4F81BD}
\definecolor{rred}{HTML}{C0504D}
\definecolor{ggreen}{HTML}{9BBB59}
\definecolor{ppurple}{HTML}{9F4C7C}
\begin{figure*}[!t] 
  \centering 
  \begin{tikzpicture}
    \begin{axis}[
        width  = 0.51*\textwidth,
        height = 6cm,
        major x tick style = transparent,
        ybar,
        bar width=1pt,
        legend style={at={(0.5,1.31)},anchor=north,font=\tiny},
        ymajorgrids = true,
        ylabel = {{phoneme error rate (PER) / \%}},
        ylabel style={yshift=-6pt},
        ymin=0.0,
        ymax=62.0,
        xlabel = {{supervised minutes (paired fraction $\gamma$)}},
        xticklabel style = {font=\tiny,yshift=1ex},
        symbolic x coords={
          5 ($\gamma{=}0.0005$),
          10 ($\gamma{=}0.001$),
          20 ($\gamma{=}0.002$),
          50 ($\gamma{=}0.005$),
          100 ($\gamma{=}0.01$),
          199 ($\gamma{=}0.02$),
          498 ($\gamma{=}0.05$),
          996 ($\gamma{=}0.1$),
          9960 ($\gamma{=}1$),
        },
        xticklabel style={rotate=30},
        xtick = data,
        scaled y ticks = false,
    ]
    , 1.1227
        \addplot[style={bblue,fill=bblue,mark=none}]
            coordinates {
              (5 ($\gamma{=}0.0005$), 49.56) 
              (10 ($\gamma{=}0.001$), 34.23) 
              (20 ($\gamma{=}0.002$), 18.81) 
              (50 ($\gamma{=}0.005$), 10.96) 
              (100 ($\gamma{=}0.01$), 6.82) 
              (199 ($\gamma{=}0.02$), 5.61)
              (498 ($\gamma{=}0.05$), 3.30)
              (996 ($\gamma{=}0.1$), 2.49)
              (9960 ($\gamma{=}1$), 1.30)
              };
              
        \addplot[style={rred,fill=rred,mark=none}]
            coordinates {
              (5 ($\gamma{=}0.0005$), 9.36)
              (10 ($\gamma{=}0.001$), 7.22)
              (20 ($\gamma{=}0.002$), 5.55)
              (50 ($\gamma{=}0.005$), 3.64)
              (100 ($\gamma{=}0.01$), 3.48)
              (199 ($\gamma{=}0.02$), 2.57)
              (498 ($\gamma{=}0.05$), 2.32)
              (996 ($\gamma{=}0.1$), 2.27)
              };
              

        \legend{$\pp(p | g)$ [supervised-only],$\pp(g{,}\, p) = \pp(g) \pp(p | g)$ [semi-supervised]}
    \end{axis}
  \end{tikzpicture}
  \begin{tikzpicture}
    \begin{axis}[
        width  = 0.51*\textwidth,
        height = 6cm,
        major x tick style = transparent,
        ybar,
        bar width=1pt,
        legend style={at={(0.5,1.31)},anchor=north,font=\tiny},
        ymajorgrids = true,
        ylabel = {{character error rate (CER) / \%}},
        ylabel style={yshift=-6pt},
        ymin=0.0,
        ymax=62.0,
        xlabel = {{supervised minutes (paired fraction $\gamma$)}},
        xticklabel style = {font=\tiny,yshift=1ex},
        symbolic x coords={
          5 ($\gamma{=}0.0005$),
          10 ($\gamma{=}0.001$),
          20 ($\gamma{=}0.002$),
          50 ($\gamma{=}0.005$),
          100 ($\gamma{=}0.01$),
          199 ($\gamma{=}0.02$),
          498 ($\gamma{=}0.05$),
          996 ($\gamma{=}0.1$),
          9960 ($\gamma{=}1$),
        },
        xticklabel style={rotate=30},
        xtick = data,
        scaled y ticks = false,
    ]
    , 1.1227
        \addplot[style={bblue,fill=bblue,mark=none}]
            coordinates {
              (5 ($\gamma{=}0.0005$), 60.64) 
              (10 ($\gamma{=}0.001$), 45.65) 
              (20 ($\gamma{=}0.002$), 20.18) 
              (50 ($\gamma{=}0.005$), 13.48) 
              (100 ($\gamma{=}0.01$), 10.52) 
              (199 ($\gamma{=}0.02$), 10.02)
              (498 ($\gamma{=}0.05$), 6.56)
              (996 ($\gamma{=}0.1$), 3.92)
              (9960 ($\gamma{=}1$), 1.73)
              };
              
        \addplot[style={rred,fill=rred,mark=none}]
            coordinates {
              (5 ($\gamma{=}0.0005$), 9.47)
              (10 ($\gamma{=}0.001$), 4.54)
              (20 ($\gamma{=}0.002$), 4.23)
              (50 ($\gamma{=}0.005$), 3.76)
              (100 ($\gamma{=}0.01$), 3.18)
              (199 ($\gamma{=}0.02$), 2.93)
              (498 ($\gamma{=}0.05$), 3.62)
              (996 ($\gamma{=}0.1$), 2.84)
              };
              

        \legend{$\pp(g | p)$ [supervised-only],$\pp(p{,}\, g) = \pp(p) \pp(g | p)$ [semi-supervised]}
    \end{axis}
  \end{tikzpicture}

  \caption{Phoneme ($p$) and grapheme ($g$) prediction error rates at varying paired fractions ($\gamma$), for a reference supervised-only model (blue) and the proposed generative model consuming both paired and unpaired data (red).}
  \label{fig:results}
\end{figure*}
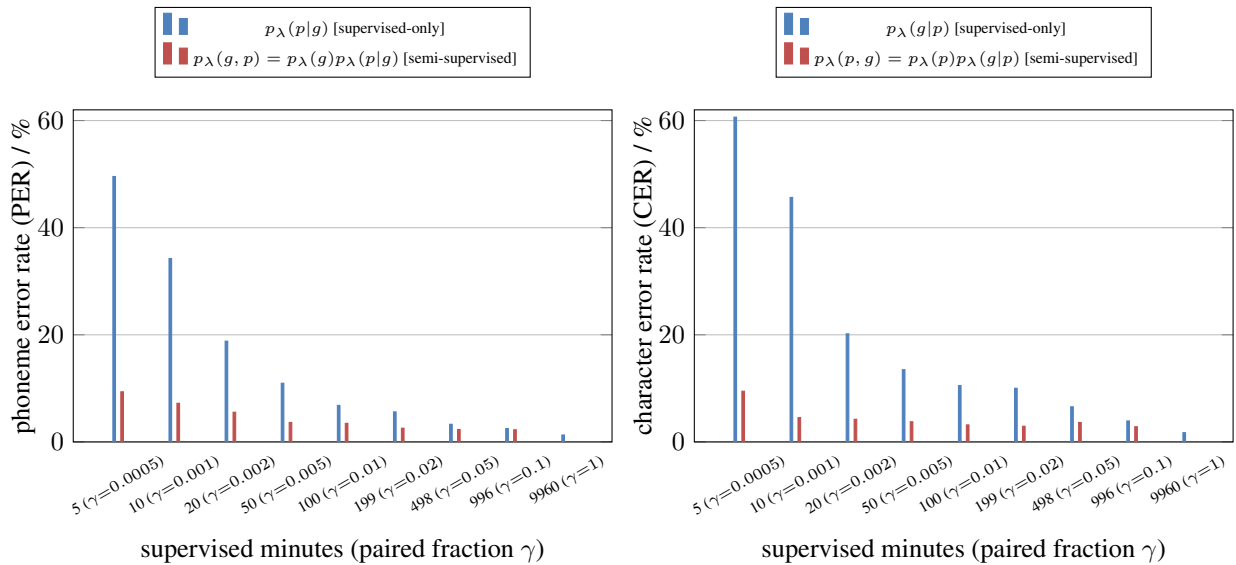 
\begin{table*}[t]
  \footnotesize
  \centering
  \begin{tabular}{l c c c c c c c c}
    \toprule
    \multirow{5}{*}{Ablation condition} & \multicolumn{8}{c}{supervised minutes (paired fraction $\gamma$)} \\
    \cmidrule(r){2-9}
     & \multicolumn{2}{C{1.5cm}}{5 min ($\gamma{=}0.0005$)} & \multicolumn{2}{C{1.5cm}}{10 min ($\gamma{=}0.001$)} & \multicolumn{2}{C{1.5cm}}{20 min ($\gamma{=}0.002$)} &  \multicolumn{2}{C{1.5cm}}{50 min ($\gamma{=}0.005$)}  \\
    \cmidrule(r){2-3} \cmidrule(r){4-5} \cmidrule(r){6-7} \cmidrule(r){8-9}
    & PER & CER & PER & CER & PER & CER & PER & CER \\
    \midrule
    Only supervised & 49.6 & 60.6 & 34.2 & 45.7 & 18.8 & 20.2 & 11.0 & 13.5 \\
    \addlinespace
    Proposed (semi-supervised) & \textbf{9.4} & \textbf{9.5} & \textbf{7.2} & \textbf{4.5} & 5.6 &
      4.2 & \textbf{3.6} & \textbf{3.8} \\
    \addlinespace
    $y_t$ w/ receptive field to $y_{t-1}$ & 64.7 & 32.5 & 20.3 & 20.5 & 9.6 & 10.5 & 7.6 & 10.2  \\
    $y_t$ w/ receptive field to $y_{<t}$ & 55.6 & 48.5 & 13.0 & 12.3 & 7.5 & 10.8 & 4.1 & 4.3 \\
    \addlinespace
    no $\pp(x, y)$ pre-training & 121.2 & 187.4 & 114.1 & 204.8 & 112.1 & 216.0 & 119.2 & 210.5 \\
    \addlinespace
    $T$ = 0.8 & 73.2 & 217.0 & 70.5 & 5.6 & \textbf{3.4} & \textbf{3.8} & \textbf{3.6} & 4.1 \\
    $T$ = 1 & 111.5 & 196.6 & 112.6 & 217.3 & 11.6 & 187.8 & 8.8 & 5.9\\
    \addlinespace
    update $\pp(x)$ with ELBO & 100.6 & 173.8 & 63.3 & 18.3 & 9.7 & 7.0 & 3.9 & 4.4 \\
    \addlinespace
    \bottomrule
  \end{tabular}
  \caption{Ablation studies at various supervision rates. \emph{Proposed} corresponds to $T=0.5$, $y_t$ w/o receptive field to $y_{<t}$, pre-trained $\pp(x, y)$, and no update of $\pp(x)$ with ELBO.}
  \label{tab:ablations}
\end{table*}

\subsection{Breaking a substitution cipher}
\label{sec:cipher-experiments}
We first apply the proposed approach to the task of breaking a substitution cipher as outlined in
\sref{sec:ident-example}.
In this case the marginal $\pp(y)$ is fully tractable and we do not have to use a variational approximation.
This investigates the ability of the proposed approach to learn from no paired data.

The experimental setup is as follows.
Our training data is a randomly chosen subset of $2\,000$ utterances from the LibriTTS corpus 
\citep{zen2019libritts}.
For each example, we derive plaintext $x$ by lowercasing and removing punctuation from the text transcript
and pass this through a fixed permutation of the $27$-character grapheme alphabet to obtain ciphertext $y$.
This yields a total of roughly $140\,000$ grapheme tokens.
We set $\gamma = 0$ and $\alpha = \beta = \half$, using half the data for collecting bigram frequencies $B_{i j}$
on the plaintext and half for collecting bigram frequencies $C_{p q}$ on the ciphertext.
Length $2$ statistics appear to be sufficient for this task.
We learn the observation matrix $O \in \reals^{27 \times 27}$, parameterized in terms of its logits using
a softmax for each row to ensure that $O$ is stochastic, to minimize the bigram loss
$l = \sum_{p, q} C_{p q} \log([\tr{O} B O]_{p q})$.

Training finds the correct mapping $O$ between plaintext and ciphertext for roughly $80\%$ of training runs.
Training success is very binary, typically either succeeding essentially perfectly with near-zero $\KLD{\pt(y)}{\pp(y)}$ and decoding error rate, or failing with a large KL divergence value and error rate.
This suggests that the training loss can potentially serve as an indicator of when training fails.
These results demonstrate that learning from zero paired data is possible with the proposed approach, as well
as highlighting the substantial challenge that local optima present in this regime.
Further experimental details and results are given in \sref{sec:cipher-experiments-more}.

\subsection{Learning to relate spoken and written language}
We apply the proposed approach to learning the relationship between spoken and written language from little paired data. We represent spoken language using phoneme sequences derived from audio. We model the joint distribution of the grapheme sequence and phoneme sequence and evaluate the model on utterance-level \emph{grapheme-to-phoneme (g2p)} and \emph{phoneme-to-grapheme (p2g)} tasks.

\subsubsection{Experimental setup}
\label{sec:setup}
Our experimental setup is as follows.
We use LibriTTS \citep{zen2019libritts}.
The phoneme sequence for each utterance is obtained by forced alignment using the decoder graph of possible verbalizations and pronunciations of the text transcript, discarding any timing information.
We limit training to utterances with grapheme and phoneme sequences of at most $96$ tokens, yielding $166$ hours of data.
We randomly partition the training set based on a hash of the string-valued utterance key, selecting a fraction $\gamma$ of utterances as our paired speech--text examples and evenly splitting the rest into text-only and speech-only datasets, that is $\alpha=\beta=\frac{1 - \gamma}{2}$.
Output distribution temperature (described in \sref{sec:training-summary}) $T = 0.5$ is used during training. Decoding is performed by stochastic sampling from $\pp(y | x)$, also with $T = 0.5$. We pre-train (described in \sref{sec:training-summary}) for $100\,000$ steps (though far fewer steps typically suffice). We compute \emph{phoneme error rate (PER)} for a generative model with $x$ a grapheme sequence and $y$ a phoneme sequence, and \emph{character error rate (CER)} for a generative model with $x$ a phoneme sequence and $y$ a grapheme sequence. To compute CER, each grapheme sequence is normalized by lowercasing and removing punctuation.

The prior, decoder and variational posterior are all modeled autoregressively. We use a recurrent neural net (RNN)-based architecture for the prior. The decoder and variational posterior are each parameterized as a seq2seq model with monotonic attention, similar to \emph{listen, attend and spell (LAS)} \citep{chan2016listen} and Tacotron \citep{wang2017tacotron}. To impose time locality on the decoder $\pp(y | x)$ as discussed in \sref{sec:ident}, the input to the final RNN predicting the distribution over $y_t$ consists only of glimpses of $x$ chosen by the attention mechanism and does not directly include any information about $y_{<t}$. No such constraint is imposed on the variational posterior $\qq(x | y)$ to ensure it remains as flexible as possible. Detailed model architectures are given in \sref{app:sec:arch}. 

\subsubsection{Experimental results}
\figref{fig:results} shows the phoneme and character error rates for utterance-level g2p and p2g on the LibriTTS test set at various paired supervision levels ($\gamma$ from $0.0005$ to $1$), both for a reference model trained only on the available paired data (blue bars) and for the proposed semi-supervised generative model (red bars). We can see that the proposed approach is able to make effective use of unpaired data to improve its predictions. Even very small amounts of paired data ($5$ minutes) are sufficient to effectively learn the association between spoken and written language. Numeric values are given in \tabref{tab:results-sup} and \tabref{tab:results-semisup} in \sref{app:sec:exp}.

\subsubsection{Initialization}
When approaching $\gamma=0$ (no paired data), the model exhibited more sensitivity to random initializations and $\gamma=0.0002$ ($2$ minutes of paired data) was the cutoff at which the model was never able to recover from poor local optima. This aligns with local optima issue discussed in \sref{sec:ident}. To investigate the effect of initialization, we pre-trained the decoder $\pp(y | x)$ with a small amount of paired data ($50$ minutes) and then continued training using only unpaired data. This gave metric improvements similar to the results reported in \figref{fig:results}. This suggests that fully unpaired training ($\gamma=0$) could be attainable with better initialization or optimization methods.

\subsubsection{Ablations}
\tabref{tab:ablations} lists several ablation studies.
When predicting $y_t$, making the decoder more powerful by allowing a receptive field to $y_{<t}$, either to $y_{t-1}$ only, or to a summary of $y_{<t}$ provided by the attention RNN state (see \sref{app:sec:arch}) degrades the performance particularly at very low supervision levels. Note that the weaker proposed decoder imposes stricter time locality and the results support our hypothesis on the importance of time locality for recovering the joint from unpaired samples (see \sref{sec:ident}). Pre-training the generative model was crucial to obtaining well-behaved training dynamics. Sampling at full temperature ($T=1$) demonstrates the issue discussed in \sref{sec:training-summary} with samples from KL-trained models and shows that lowering the temperature is an effective remedy. This issue is a particular problem at very low paired fractions, presumably due to the increased reliance on accurate encoder samples when learning from $y$-only data. Finally, the prediction metrics worsens when we update the prior weights using the ELBO as hypothesized in \sref{sec:training-summary}.

\subsubsection{Samples}
\begin{table*}[t]
  \centering
\begin{tabular}{l l} 
 \toprule
 Sequence type & Sequence value \\
 \midrule
 \multirow{2}{*}{input phonemes} &
   \texttt{\small /sil \textbf{b V k m V l @ g @ n} sil w A: k @ N f O: r\textbackslash{}} \\
 & \texttt{\small \quad w @\`{} d @ g E n sil r\textbackslash{} eI z d h I z h \{ n dz sil/} \\
 \midrule
 ground truth graphemes & \textit{\textbf{Buck Mulligan}, walking forward again, raised his hands.} \\ 
 supervised-only prediction & \textit{\textbf{Buc-mullgaan}, walking forward again, raised his hands.}\\ 
 semi-supervised prediction & \textit{\textbf{Buck Mulligan}, walking forward again, raised his hands.}\\ 
 \bottomrule
\end{tabular}
\caption{Sample of an utterance-level p2g prediction.}
\label{tab:sample}
\end{table*}
Here we present a sample of the model for utterance-level p2g task in \tabref{tab:sample}. In this example \texttt{Buck Mulligan} is a rare proper noun that appears both in text-only and speech-only samples, but it is never observed in paired data. Therefore, the supervised-only model is unable to predict it correctly. However the proposed model is able to correctly associate between its phoneme and grapheme representations, although it is never presented with the pair.

\section{RELATED WORK}

In the past few years, remarkable progress has been made in supervised ASR \citep{gulati2020conformer,han2020contextnet,chan2021speechstew} and TTS \citep{wang2017tacotron,shen2018natural,ren2020fastspeech} systems, powered by availability of massive parallel text and speech corpora, like LibriSpeech \citep{panayotov2015librispeech} and LibriTTS \citep{zen2019libritts}. Due to scarcity of such resources across all languages, there has been a great interest in leveraging non-parallel data (i.e., unspoken text and untranscribed speech) which are readily available at larger scales, without the need for manual transcription.

Toward this goal, self-supervision with various self-consistency training objectives has been proven to be an effective way to pre-train speech encoder for ASR (e.g., CPC \citep{oord2018representation}, wav2vec \citep{schneider2019wav2vec}, vq-wav2vec \citep{baevski2019vq}, wav2vec 2.0 \citep{baevski2020wav2vec}, HuBERT \citep{hsu2021hubert}, W2v-BERT \citep{chung2021w2v}), and text / phoneme encoder for TTS (e.g., \citep{hayashi2019pre}, PnG BERT \citep{jia2021png}). But, these pre-trained models need to be fine-tuned with parallel data to tailor the representations to the task of interest (e.g., paralinguistics \citep{shor2022universal}, speaker verification \citep{chen2022large}, or ASR \citep{hsu2021hubert}). Semi-supervised training is an alternative approach for extracting useful information from both speech-only and text-only datasets for ASR \citep{li2019semi,chen2021semi} and TTS \citep{chung2019semi}.

The shared information between the two modalities, referred to as the ``common form'' in \citet{taylor2009text}, comprises a sequence of words, giving rise to variations of graphemic symbols in written form (e.g., Dr.\ versus doctor) and also variations in spoken form (e.g., prosodic inflections).

The recent developments in textless NLP models aims at uncovering parts of this connection from speech-only data \citep{lakhotia2021generative, borsos2022audiolm}. Learning to associate the spoken and written language with little or no parallel data has been extensively explored but remains unsolved. At a high level we categorize the current approaches into three groups. The first family of models use weight sharing in a multi-task speech / text representation learning setup to encourage alignment between the modalities in the encoded representation, which is always accompanied by some form of supervised objective on paired data to force that alignment (e.g., SpeechT5 \citep{ao2021speecht5}, SLAM \citep{bapna2021slam}, MAESTRO \citep{chen2022maestro}). The second family are primarily based on the back translation technique first adopted for machine translation \citep{sennrich2015improving} to incorporate monolingual data in a target language, by pairing them with automatic translation to the source language to generate synthetic paired data. This is closely related to speech chain theory that hypothesizes a connection between speech perception and production with a reinforcing feedback loop \citep{denes1993speech}. This connection has motivated the joint training of ASR / TTS models in an auto-encoding setting on non-parallel data \citep{tjandra2017listening}, with the limitation of not being able to the back-propagate to the ASR model when speech is auto-encoded, due to non-differentiable ASR textual outputs, which was later remedied by a straight-through gradient estimation \citep{tjandra2019end}. To avoid the mismatches between synthetic and real speech a similar cycle-consistency approach has been adopted on output of an ASR encoder \citep{hayashi2018back, hori2019cycle}. The use of pre-trained TTS as data augmentation \citep{rosenberg2019speech} or to guide self-supervised speech representation \citep{chen2021injecting} can also be considered variations of this technique. All of the models under this category are also trained with some amount of paired data. The third family of models is based on distribution matching, minimizing some divergence when mapping unpaired data to the other modality, including adversarially matching the phoneme distribution output by a speech encoder and phonemes derived from a text corpus \citep{liu2018completely, baevski2021unsupervised, liu2022towards} to build unsupervised ASR models. The tokens discovered by \citet{baevski2021unsupervised} have also been used as conditioning input to build an unsupervised TTS model \citep{liu2022simple}. The distribution matching family is the most principled approach and is the only one that has some success in zero paired data setup to the best of our knowledge. It is interesting to note that \citet{baevski2021unsupervised} also impose a strict time locality assumption (successive phonemes output by the generator network are assumed to be conditionally independent given the speech waveform). We discussed the potential importance of this assumption in \sref{sec:ident}.

\section{DISCUSSION}
We formulated the problem of utilizing unpaired data in seq2seq problems as a joint identifiability problem by only observing unpaired samples from the marginals, with little or no paired data, and presented a generative modeling approach to this problem, a form of distribution matching. The objective function is simply the maximum likelihood, or its lower bound (ELBO), without any extra ad hoc losses. Interestingly, the ELBO objective in the $\beta$ term of \Eqref{eq:loss-generative} results in an optimization solution with some similarity to the back translation technique, but with key differences. For instance \citet{tjandra2019end} explore several decoding strategies including greedy and beam search for back translation and our formulation suggests stochastic sampling as the principled decoding strategy (the second term in \Eqref{eq:loss-generative}).
Also, the variational noisy channel formulation in \Eqref{eq:klubo} yields both direct optimization with reverse KL divergence to the generative distribution with reparameterized samples, when output is fixed length and continuous (e.g., spectrograms), and also KL encoder loss described in \ref{sec:kl-encoder-loss}.
Furthermore, this work highlights time locality as the key to do so, which can guide the architecture choices. The experimental results in \figref{fig:results} show that even tiny amount ($5$ minutes) of paired data is sufficient to enable learning the association between unpaired data.

Accurate density modeling is crucial for this type of generative modeling. For instance uncalibrated $\log \pp(y|x)$ can adversely affect the balance in the second (ELBO) term of $l^\text{gen}$. To examine the potential of the proposed approach without such concerns, we have only experimented with phoneme and grapheme with categorical distribution, which is universal. This is not a common task to be able to do comparisons with other methods. But, it served as nice test bed for this novel approach. A natural question is how these results hold when we move to end-to-end ASR and TTS, for which we would need accurate speech density estimation.
Alternatively, given the success of tokenized speech representation a natural extension of this work is to use such representations \citep{lakhotia2021generative,zeghidour2021soundstream}. Also, the masked language model (MLM)-based pre-trained models under uniform masking regime \citep{ghazvininejad2019mask} can be interpreted as probability distributions estimated with maximum likelihood, and hence it can be directly plugged into our formulation, either as $\pp(x)$ or $\pt(y)$ in \Eqref{eq:loss-generative} and \Eqref{eq:loss-variational}, making this generative modeling approach also nicely amenable to the commonly adopted pre-training / fine-tuning workflow.

Finally, similar conditional independence structure appears in other problems (e.g., machine translation) and the approach and discussions here could be more widely applicable.

\bibliography{paper}
\bibliographystyle{apalike}

\clearpage
\newpage
\appendix
\onecolumn
\aistatstitle{Learning the joint distribution of two sequences using little or no paired data: \\
supplementary material}

\section{IDENTIFIABILITY: DETAILS}
\label{sec:ident-more}
In this section we give more details about identifiability following on from the discussion in \sref{sec:ident}.

We first give details about the singular value decomposition-based approach when $O$ is a permutation matrix,
corresponding to a substitution cipher.
Compute the singular value decomposition of the plaintext bigram frequencies $B$ and the ciphertext bigram
frequencies $C$ as
\begin{align}
  B &= U_x \Lam_x \tr{V_x}
\\
  C &= U_y \Lam_y \tr{V_y}
\end{align}
where $U_x, V_x, U_y, V_y$ are real orthogonal matrices and $\Lam_x$ and $\Lam_y$ are real diagonal matrices with
entries increasing along the diagonal.
We know that $C$ must also equal $\tr{O} B O = (\tr{O} U_x) \Lam_x \tr{(\tr{O} V_x)}$ which is also a
singular value decomposition of $C$.
Standard results on the uniqueness of the singular value decomposition, which can be obtained by
considering the eigenvalues and eigenvectors of $C \tr{C}$ and $\tr{C} C$, show that $\Lam_x = \Lam_y = \Lam$
and that, if the singular values are all distinct and non-zero, then the left and right singular vectors are 
determined up to sign, that is $U_y = \tr{O} U_x S$ and $V_y = \tr{O} V_x T$ for two diagonal matrices $S$ and $T$
with $1$s or $-1$s along their diagonal.
Thus $O = U_x S \tr{U_y} = V_x T \tr{V_y}$. 
Since $O \ones = \ones$, we have $S \tr{U_y} \ones = \tr{U_x} \ones$, which allows us to recover $S$, and
similarly $T$.
Thus $O$ is identifiable given no paired data in this case as long as the singular values are distinct.

Bigram binary identifiability details.
We wish to show that if $B$ is full rank and $b_0 \neq \half$ then we have identifiability given no paired data.
We first consider sequences of length two.
We suppose that $\tr{O} B O = \tr{\Ot} B \Ot$ for some $O$, $\Ot$ and ask whether
this implies $O = \Ot$.
The expression $\tr{O} B O$ is a quadratic function of $\eta$ and $\zeta$.
By explicitly expanding in terms of $\eta$ and $\zeta$, it may be verified that the only possible solutions are
\begin{equation}
  \label{eq:o-bigram-binary-quadratic}
  O = \left[
    \begin{matrix}
      c_0 & c_1
      \\
      c_0 & c_1
    \end{matrix}
  \right] \pm \det(\Ot)
  \left[
    \begin{matrix}
      b_1 & -b_1
      \\
      -b_0 & b_0
    \end{matrix}
  \right]
\end{equation}
where the determinant $\det(\Ot)$ is also equal to $\det(C) / \det(B)$.
In some cases one of these solutions will have a negative entry and so not represent a valid decoder,
in which case we have identifiability.
This is more likely when $\Ot$ has large determinant.
As before we also technically have identifiability in the uninteresting case where $\eta = \zeta$.
Thus in general based on sequences of length two alone we cannot uniquely identify $O$.
Now consider sequences of length three.
Define the trigram distributions
\begin{align}
  B_{i j k} &= \pp(x_0 = i, x_1 = j, x_2 = k)
  \\
  C_{p q r} &= \pt(y_0 = p, y_1 = q, y_2 = r)
  \\
  \label{eq:trigram-binary-d}
  D_{p q r} &= \pp(y_0 = p, y_1 = q, y_2 = r) = \sum_{i, j, k} B_{i j k} O_{i p} O_{j q} O_{k r}
\end{align}
We suppose that
$\sum_{i, j, k} B_{i j k} O_{i p} O_{j q} O_{k r} = \sum_{i, j, k} B_{i j k} (\Ot)_{i p} (\Ot)_{j q} (\Ot)_{k r}$
for some $O$, $\Ot$ and ask whether this implies $O = \Ot$.
We only need to worry about distinguishing the two solutions in \Eqref{eq:o-bigram-binary-quadratic}.
By plugging in these two solutions into \Eqref{eq:trigram-binary-d} it can be shown that if
\begin{equation}
  \label{eq:trigram-binary-condition}
  \sum_{i, j, k} B_{i j k} (b^\perp)_i (b^\perp)_j (b^\perp)_k \neq 0
\end{equation}
where $b^\perp = [-b_1, b_0]$ then $O = \Ot$.
The above holds for general prior distributions $\pp(x)$.
If $\pp(x)$ is Markovian then $B_{i j k} = B_{i j} B_{j k} / b_j$ and the condition
\Eqref{eq:trigram-binary-condition} is equivalent to $b_0 \neq \half$.
Thus if $B$ is full rank and $b_0 \neq \half$ then we have identifiability given no paired data.
This holds for any choice of $O$ specifying the decoder.

It is also interesting to consider the form of loss landscape under \Eqref{eq:loss-decomp} in the case
of a Markovian prior and time-independent and time-synchronous decoder.
In the case where the $x$ and $y$ alphabets both have size two and $B$ is full rank, it can be shown
that there are at most three stationary points, one saddle point corresponding to the
\emph{unigram initialization} $O = \ones \tr{c}$ which has the correct unigram marginals for $y$
but which ignores $x$, and up to two local minima related to the two solutions discussed in the
identifiability discussion above.
We examined the case of larger alphabets experimentally in \sref{sec:experiments}.

\section{EXPERIMENTS: DETAILS}

\subsection{Breaking a substitution cipher: details}
\label{sec:cipher-experiments-more}
\begin{figure}[t]
  \centering
  \includegraphics[trim={2cm 0cm 2cm 0cm},width=\textwidth,clip]{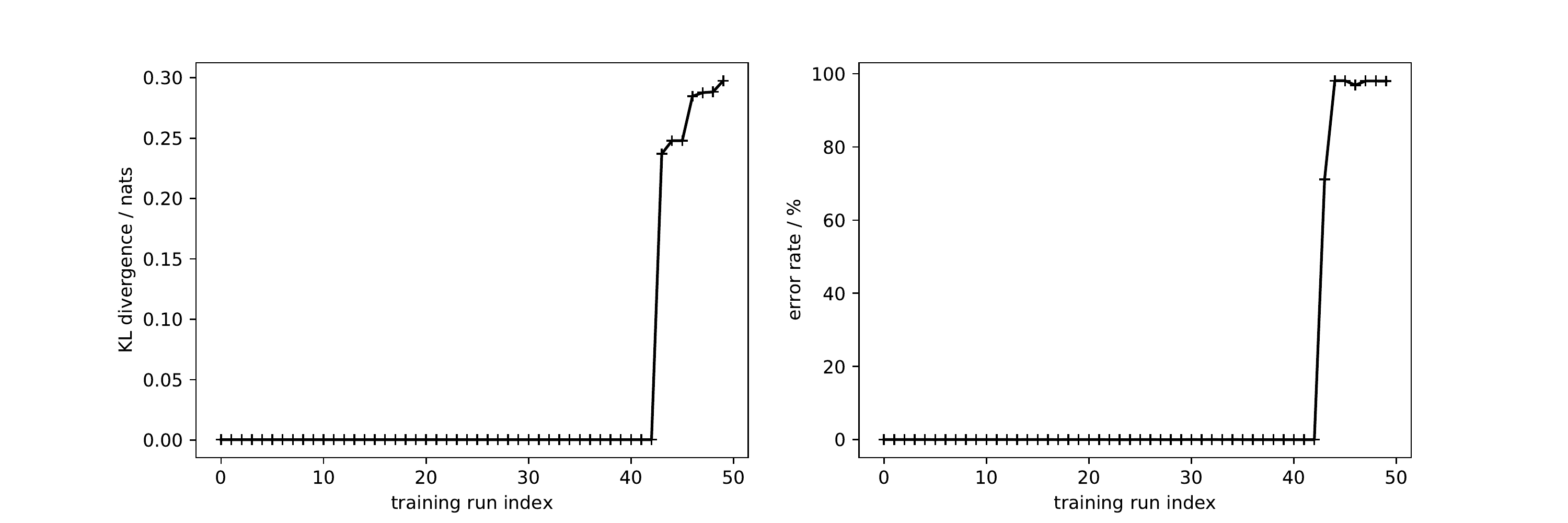}
  \caption{%
    The performance of the learned models across $50$ random training runs sorted by increasing final training loss.
    Performance is measured by the training loss, which is the bigram KL divergence
    $\KLD{\pt(y_0, y_1)}{\pp(y_0, y_1)}$, and the decoding error rate.
    Training success is very binary and there is a strong correlation between training loss and error rate.
  }
  \label{fig:cipher-runs-kl-err}
\end{figure}
In this section we give further details of the experiments breaking a substitution cipher described in
\sref{sec:cipher-experiments}.
To initialize $O$, we take the \emph{unigram initialization} $O = \ones \tr{c}$, where $c$ is the observed
ciphertext unigram frequencies, and apply a small random normal perturbation to the logits to break symmetry
since the unigram initialization is a stationary point.
For optimization we use an initial phase of stochastic gradient descent with learning rate $10$ to get in roughly
the right region of parameter space followed by Adam \citep{kingma2014adam} with learning rate $0.01$ to converge
to the precise local optimum, since we find this works substantially better than either optimizer on its own.
To investigate the effect of random initialization, we perform $50$ training runs, sort by the achieved training
loss, and plot the training loss $\KLD{\pt(y_0, y_1)}{\pp(y_0, y_1)}$ and decoding error rate in
\figref{fig:cipher-runs-kl-err}.
The binary nature of success or failure mentioned in \sref{sec:cipher-experiments} is clearly visible.
This is somewhat encouraging since it suggests that the training loss can potentially serve as a reliable
detector of when to discard a training run, at least when the generative model matches the true generating
process well as here.

\begin{figure}[t]
  \centering 
  \includegraphics[trim={4cm 0 4cm 1cm},scale=0.42,clip]{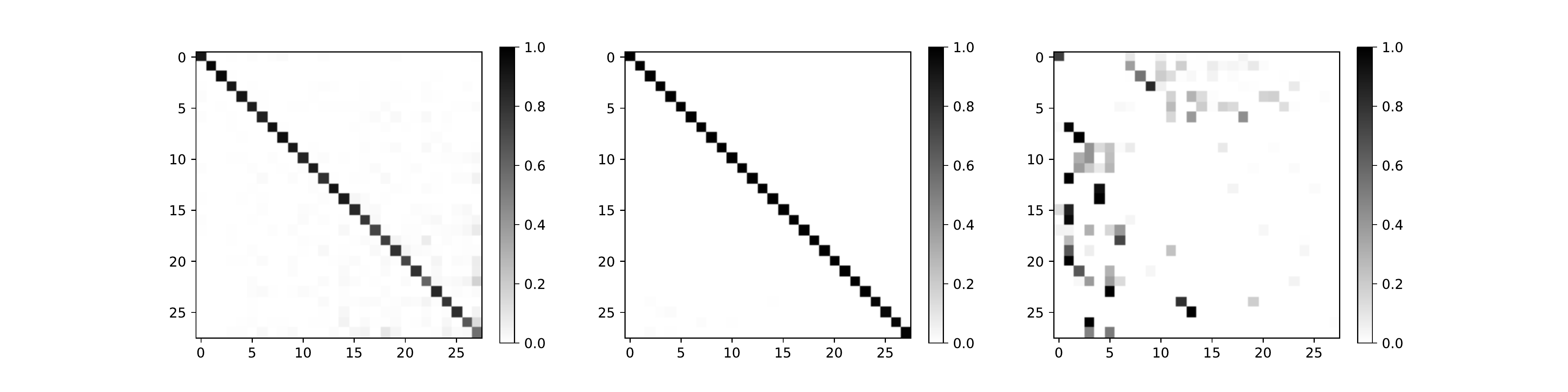}
  \caption{%
    Examples of learned $O$ for singular value decomposition (left), a successful gradient descent run (middle) and
    a failed gradient descent run (right).
    The vertical axis is the plaintext grapheme $x_t$ and the horizontal axis is the ciphertext grapheme
    $y_t$ (with columns reordered so that the true mapping is the identity).
    Thus each row represents the learned probability distribution $\pp(y_t | x_t)$ for a particular $x_t$.
    The SVD solution is not perfect due to its sensitivity to the plaintext and ciphertext coming from distinct
    utterances.
    The failed run finds a suboptimal local minimum in the training loss, and essentially the same
    $O$ is found by multiple different failed training runs.
    In this case the failed run appears to have learned to map vowels to consonants and consonants to vowels.
  }
  \label{fig:cipher-o-success-failure}
\end{figure}
Examples of the $O$ learned when training succeeds and fails are shown in \figref{fig:cipher-o-success-failure}.
The symbol table used is
\begin{align*}
  &\texttt{0000000000111111111122222222}
  \\
  &\texttt{0123456789012345678901234567}
  \\
  &\texttt{\^{}\textvisiblespace{}eaoiutnhsrdlmcwfygpbvkxqjz}
\end{align*}
where \texttt{\^} is a start-of-sequence and end-of-sequence symbol and \text{\textvisiblespace} is the space
character.
Note the grouping of vowels then consonants.
\begin{figure}[t]
  \centering
  \includegraphics[trim={5cm 2cm 4cm 1cm},width=\textwidth,clip]{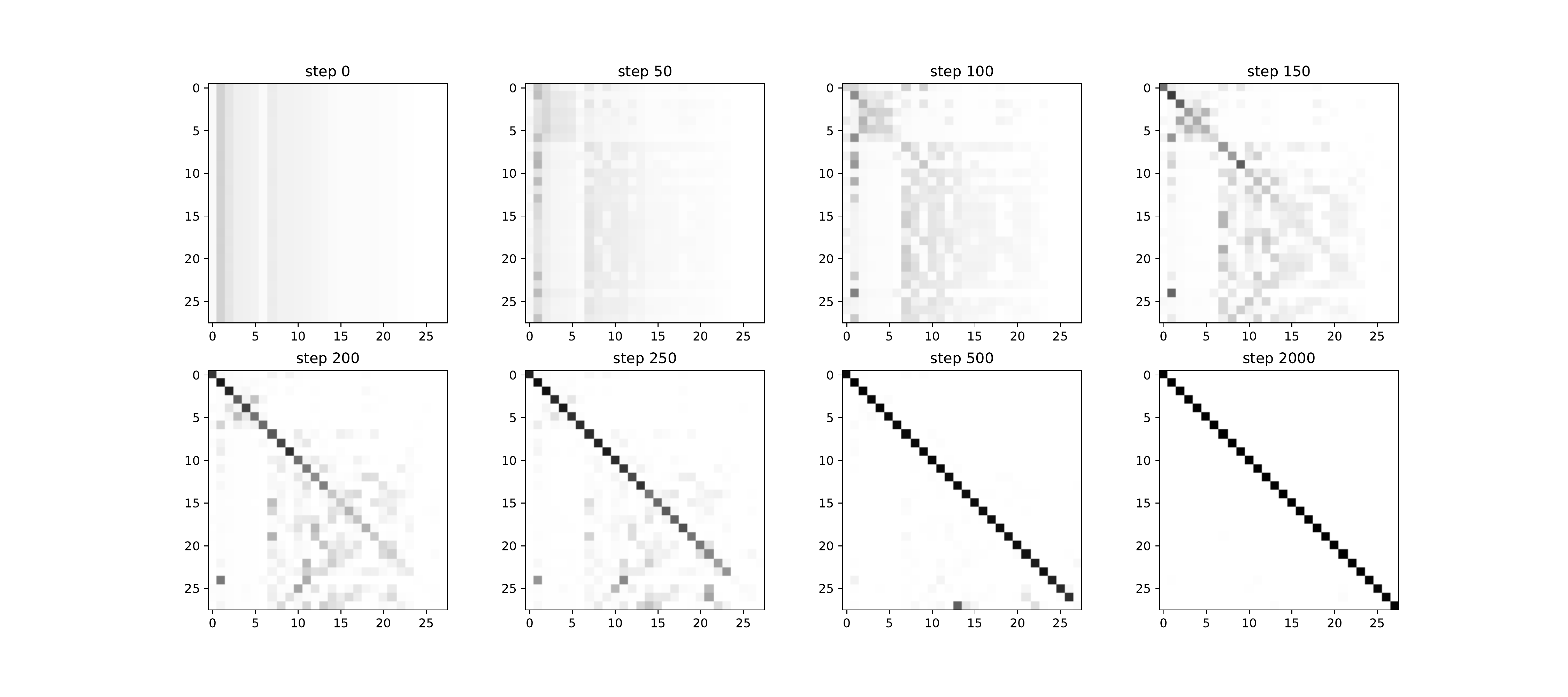}
  \caption{%
    The matrix $O$ learned after various numbers of training steps for a successful training run.
    By step $50$, a tentative mapping of vowels to vowels and consonants to consonants has been learned.
    By step $100$, the correspondence for the space character and a few of the most common individual vowels
    and consonants has been tentatively, and by step $150$ clearly, learned.
    Clear knowledge of a few symbols opens up contexts for learning the association of other symbols
    based on their statistical properties, and learning progresses rapidly for less and less common symbols
    through steps $200$ and $250$.
    Finalizing the precisely correct association for very rare consonants such as \texttt{z} takes many steps.
  }
  \label{fig:cipher-movie-frames}
\end{figure}
The matrix $O$ learned at various stages of training for a successful training run is shown in 
\figref{fig:cipher-movie-frames}.
For coherence in this case Adam was used throughout.
This shows how the association for common symbols is learned earlier in training, potentially unlocking
additional known contexts in which to learn rarer symbols.

\subsection{Learning to relate spoken and written language: details}

\subsubsection{Model architecture and training hyperparameters}
\label{app:architecture}
\label{app:sec:arch}

\paragraph{Prior}
\begin{table}[t]
    \centering
    \begin{tabular}{ll}
    \toprule
    Module & Hyperparameters    \\
    \midrule
     One-hot \\
     Causal Conv1D   &  filters = 128, kernel size = 8, activation = ReLU \\
     RNN w/ GRU cells &  units = 128, zoneout probability = 0.01 \\
     Dropout  &  dropout rate = 0.05 \\
     Dense   &  units = 128, activation = ReLU\\
    \bottomrule
    \end{tabular}
    \caption{Summary of the autoregressive prior $\pp(x)$ architecture and hyperparameters.}
    \label{tab:arc:prior}
\end{table}
The prior $\pp(x)$ is parameterized by an RNN-based autoregressive model, with a causal convolutional preprocessing of past samples. The detailed building blocks of this architecture is summarized in \tabref{tab:arc:prior}.

\paragraph{Decoder parameterization}
\begin{table}[t]
    \centering
    \begin{tabular}{ll}
    \toprule
    Module & Hyperparameters    \\
    \midrule
    Input encoder & One-hot \\
                  &  Conv1D: filters = 128, kernel size = 8, activation = ReLU \\
                  &  Conv1D: filters = 128, kernel size = 8, activation = ReLU \\
    Autoregressive decoder & Causal Conv1D filters = 64, kernel size = 12, activation = ReLU \\
    & attention LSTM units = 64 \\
    & montonic GMM attention \citep{battenberg2020location} \\
    & $\rightarrow$num components = 5, num heads = 1, units = 32 \\
    & $\rightarrow$init offset bias = 1.0, init scale bias = 5.0 \\
    & decoder LSTM (units = 64)\\ 
    & $\rightarrow$input to decoder LSTM: only the attention glimpse \\
    \bottomrule
    \end{tabular}
    \caption{Summary of the autoregressive decoder $\pp(y | x)$ architecture and hyperparameters.}
    \label{tab:arch-decoder}
\end{table}
The decoder $\pp(y | x)$ architecture is adopted from the Tacotron model \citep{wang2017tacotron} with some differences. We use a stack of two 1D convolution layers as input encoder, and the recurrent GMM attention model extract glimpse of the encoded input to feed to the decoder LSTM. When predicting $y_t$ no receptive field to $y_{<t}$ is allowed. The details of this architecture is listed in \tabref{tab:arch-decoder}. Note that a more powerful decoder with receptive field to past samples did not make a difference on prediction metrics of the supervised-only model. So, for simplicity, we kept the decoder consistent between the supervised-only and semi-supervised setups.

\paragraph{Variational posterior parameterization}
\begin{table}[t]
    \centering
    \begin{tabular}{ll}
    \toprule
    Module & Hyperparameters    \\
    \midrule
    Input encoder & One-hot \\
                  &  Conv1D: filters = 128, kernel size = 8, activation = ReLU \\
                  &  Conv1D: filters = 128, kernel size = 8, activation = ReLU \\
    Autoregressive posterior & Causal Conv1D filters = 64, kernel size = 12, activation = ReLU \\
    & attention LSTM units = 64 \\
    & montonic GMM attention \citep{battenberg2020location} \\
    & $\rightarrow$num components = 5, num heads = 1, units = 64 \\
    & $\rightarrow$init offset bias = 1.0, init scale bias = 5.0 \\
    & decoder LSTM (units = 64)\\ 
    & $\rightarrow$input to decoder LSTM: attention glimpse, attention RNN state, $x_{t-1}$ \\
    \bottomrule
    \end{tabular}
    \caption{Summary of the autoregressive variational posterior $\qq(x | v)$ architecture and hyperparameters.}
    \label{tab:arch-encoder}
\end{table}
The variational posterior $\qq(x | v)$ is very similar to the decoder, but slightly more powerful by allowing prediction of $x_t$ to have receptive field to $x_{<t}$ via state of the attention LSTM, and also having a skip connection from input to decoder RNN. The details of this architecture is listed in \tabref{tab:arch-encoder}.

\paragraph{Training} We use separate Adam optimizers \citep{kingma2014adam} for generative parameters $\lam$ and variational parameters $\nu$. The generative optimizer learning rate is piecewise constant of values of $1e{-}3$, $5e{-}4$, $3e{-}4$, $1e{-}4$, $5e{-}5$,  changing every $50\,000$ steps, and fixed after $200\,000$ steps. The variational optimizer uses fixed higher learing rate of $3e{-}3$ to keep the variational posterior up to date with respect to generative parameters. Both optimizer apply gradient clipping.

\subsubsection{Experimental results: details}
\label{app:sec:exp}
\begin{table}[t]
  \footnotesize
  \centering
  \begin{tabular}{l c c c c}
    \toprule
    \multirow{2}{3.2cm}{paired minutes (fraction)} & \multicolumn{2}{C{1.9cm}}{dev} & \multicolumn{2}{C{1.9cm}}{test} \\
    \cmidrule(r){2-3} \cmidrule(r){4-5}
    & PER / \% & CER / \% & PER / \% & CER / \% \\
    \midrule
      5 ($\gamma{=}0.0005$)& 47.8 & 61.6 & 49.6 & 60.6 \\
      10 ($\gamma{=}0.001$)& 33.2 & 46.9 & 34.2 & 45.7 \\
      20 ($\gamma{=}0.002$)& 19.0 & 21.7 & 18.8 & 20.2 \\
      50 ($\gamma{=}0.005$)& 12.1 & 14.6 & 11.0 & 13.5 \\
      100 ($\gamma{=}0.01$)& 7.2 & 11.9 & 6.8 & 10.5 \\
      199 ($\gamma{=}0.02$)& 5.0 & 9.6 & 5.6 & 10.0 \\
      498 ($\gamma{=}0.05$)& 3.4 & 5.7 & 3.3 & 6.6 \\
      996 ($\gamma{=}0.1$)& 2.8 & 4.5 & 2.5 & 3.9 \\
      9960 ($\gamma{=}1$)& 0.9 & 2.1 & 1.3 & 1.7 \\
    \bottomrule
  \end{tabular}
  \caption{%
    Phoneme and character error rates for utterance-level g2p and p2g at various paired supervision rates $\gamma$ on LibriTTS dev and test sets for supervised-only approach.
  }
  \label{tab:results-sup}
\end{table}
\begin{table}[t]
  \footnotesize
  \centering
  \begin{tabular}{l c c c c}
    \toprule
    \multirow{2}{3.2cm}{paired minutes (fraction)} & \multicolumn{2}{C{1.9cm}}{dev} & \multicolumn{2}{C{1.9cm}}{test} \\
    \cmidrule(r){2-3} \cmidrule(r){4-5}
    & PER / \% & CER / \% & PER / \% & CER / \% \\
    \midrule
      5 ($\gamma{=}0.0005$)& 11.7 & 25.6 & 9.4 & 9.5 \\
      10 ($\gamma{=}0.001$)& 11.1 & 5.3 & 7.2 & 4.5 \\
      20 ($\gamma{=}0.002$)& 6.3 & 4.7 & 5.6 & 4.2 \\
      50 ($\gamma{=}0.005$)& 4.2 & 3.7 & 3.6 & 3.8 \\
      100 ($\gamma{=}0.01$)& 3.6 & 3.3 & 3.5 & 3.2 \\
      199 ($\gamma{=}0.02$)& 3.0 & 2.8 & 2.6 & 2.9 \\
      498 ($\gamma{=}0.05$)& 3.0 & 3.1 & 2.3 & 3.6 \\
      996 ($\gamma{=}0.1$)& 2.2 & 3.2 & 2.3 & 2.8 \\
    \bottomrule
  \end{tabular}
  \caption{%
    Phoneme and character error rates for utterance-level g2p and p2g at various paired supervision rates $\gamma$ on LibriTTS dev and test sets for the proposed semi-supervised generative modeling approach.
  }
  \label{tab:results-semisup}
\end{table}
The numerical values of prediction errors for supervised-only and semi-supervised models are reported in \tabref{tab:results-sup} and \tabref{tab:results-semisup} respectively.
\end{document}